# A Bi-level Framework for Traffic Accident Duration Prediction: Leveraging Weather and Road Condition Data within a Practical Optimum Pipeline

**Rafat Tabassum Sukonna[1,2] (sukonna.ipebuet@gmail.com), Soham Irtiza Swapnil[2,3,*]** (swapnil.buetbme@gmail.com)

[1] Ahsanullah University of Science and Technology, Tejgaon, Dhaka.

[2] Bangladesh University of Engineering and Technology, Dhaka, Bangladesh.

[3] Advanced Chemical Industries Limited, Tejgaon, Dhaka.

[*] Corresponding Author

*Both authors have equal contribution to this article*

## Abstract:

Due to the stochastic nature of events, predicting the duration of a traffic incident presents a formidable challenge. Accurate duration estimation can result in substantial advantages for commuters in selecting optimal routes and for traffic management personnel in addressing non-recurring congestion issues. In this study, we gathered accident duration, road conditions, and meteorological data from a database of traffic accidents to check the feasibility of a traffic accident duration pipeline without accident contextual information data like accident severity and textual description. Multiple machine learning models were employed to predict whether an accident's impact on road traffic would be of a short-term or long-term nature, and then utilizing a bimodal approach the precise duration of the incident's effect was determined. Our binary classification random forest model distinguished between short-term and long-term effects with an 83% accuracy rate, while the LightGBM regression model outperformed other machine learning regression models with Mean Average Error (MAE) values of 26.15 and 13.3 and RMSE values of 32.91 and 28.91 for short and long-term accident duration prediction, respectively. Using the optimal classification and regression model identified in the preceding section, we then construct an end-to-end pipeline to incorporate the entire process. The results of both separate and combined approaches were comparable with previous works, which shows the applicability of only using static features for predicting traffic accident duration. The SHAP value analysis identified weather conditions, wind chill and wind speed as the most influential factors in determining the duration of an accident.

## Introduction:

With the accelerated urbanization of the world, traffic congestion has escalated into an increasingly serious problem. Economically, socially, and environmentally, traffic gridlock can have significant negative effects. One of the major factors leading to traffic congestion [1] is the presence of traffic events that jeopardize safety and impede the smooth flow of vehicles, including accidents and incidents involving the release of hazardous materials. Typically, traffic obstruction is classified as either recurrent or non-recurrent [2,3]. Recurrent traffic congestion occurs when the road's capacity is exceeded, whereas non-recurrent congestion is a transitory decrease in regular capacity caused by incidents, maintenance operations or construction projects, and special gatherings with higher-than-normal peak demand [4]. Due to the stochastic nature of non-recurrent delay in both time and space domain, predicting the non-recurrent traffic delay can be challenging [5,6].To enable managers to better manage traffic incidents, further enhance traffic safety, reduce the loss caused by traffic accident congestion, and ensure the travel safety of travellers and the quality of travel services, as well as to reduce the traffic operation burden and alleviate traffic congestion, the question is of great importance to the study of accident duration[7].

It is believed that weather conditions have a significant impact on the number of traffic accidents and fatalities, with varying effects dependent on the type of route (highways, rural roads, or urban roads). Moreover, given that the weather also affects mobility, it stands to reason that the weather's effects on the frequency of injury accidents and fatalities are also influenced by concurrent changes in mobility [8]. In a previous meta-analysis [9], temperature has also been linked to traffic accidents. The public has become more aware of the significance of response time and incident management effectiveness, particularly in the context of traffic accidents. To ensure the long-term viability of traffic systems, a speedy recovery from these catastrophes [10] is crucial. Predicting accurately the duration of traffic incidents is essential for effective traffic management, improving traveler convenience and for developing an intelligent transport system.

Accident duration can be termed as the time duration between the reporting of traffic accident to the time when the effect of traffic accident on traffic flow was dismissed. This duration has been divided into four sequences: I) accident reporting time (termed as start time in US accident dataset) ii) accident response time iii) accident clearance time and iv) accident recovery time (termed as end time in the dataset) [11]. In this work we are using the time difference between the reported end time and start time provided in the dataset in minutes, which was also reported in the official traffic log. In recent years, long-term forecasting has emerged as a crucial area of research for enhancing road safety. According to previous research, traffic accidents are a type of nonrecurring traffic congestion that can reduce road capacity and increase the likelihood of a subsequent accident, resulting in increased emissions and economic loss.

In recent times, intelligent transportation systems have offered extensive assistance in transportation planning, management, public travel, and overall transportation operations. This has been made possible by the integration of sensor technology, network communication technology, data processing technology, and business applications [12,13]. Notably, machine learning (ML) modeling has introduced novel techniques for predicting accident risks and its effect on road traffic conditions, leveraging its ability to effortlessly incorporate new data sources and eliminate the linear assumptions between features and the predicted outcomes[2,14]. Several statistical modelling techniques have been utilised in the past; however, more recently, new approaches in ML modeling have emerged as a more advanced method. Artificial neural networks (ANNs)[15] , genetic algorithms [16], support vector machines (SVMs) [17], k-nearest-neighbors (kNNs) [18], and decision-trees (DTs) [19] are examples of such methods.

In a study conducted by Hamad, Alruzouq, and Zeiada et al. [20], they utilized a dataset comprising over 50 variables extracted from more than 140,000 incident records in the Houston metropolitan area of Texas. Their objective was to develop random forest (RF) models capable of predicting incident durations within the ranges of 1 to 1,440 minutes and 5 to 120 minutes. The research findings demonstrated that, in comparison to artificial neural networks (ANN), the random forest model exhibited a mean absolute error (MAE) of 14.979 minutes and 36, indicating its superior performance. In a separate study, a new framework combining machine learning with features encoded by multiple Deep Learning layers was devised to forecast the duration of an incident based on limited data. When applied to incident reports using baseline data, the method improves the accuracy of event duration prediction beyond the performance of the top ML models. Compared to conventional linear or support vector regression models, our proposed strategy can improve accuracy by 60%, and when compared to the hybrid deep learning auto encoded GBDT model, which appears to be superior to all other models, it can improve accuracy by an additional 7%. The application domain is the city of San Francisco, which contains an abundance of data on both historical traffic congestion and traffic incident records (Countrywide Traffic Accident Data set) [21]. In a separate study, using XGBoost, LightGBM, CatBoost, stacking, and elastic networks, a heterogeneous ensemble learning model was developed to predict the accident duration. The results demonstrate that not only does the model have a high degree of predictive accuracy, but it can also incorporate multiple models to provide a comprehensive measure of influence factor importance. The model's feature importance demonstrates that the accident's time, location, weather, and germane historical data are significant factors in determining its duration [7]. Another study proposed the intra-extra Joint Optimisation algorithm (IEO-ML), which extends various baseline ML models evaluated against a variety of regression situations across data sets. Consequently, the authors assessed the feature importance and demonstrated that the top 10 significant characteristics that determine how long events will last are time, location, incident type, incident reporting source, and weather. They then used incident data logs to develop a binary classification prediction approach, which allows them to classify traffic incidents as short-term or long-term. They used 40–45 minutes as the optimal dividing line between short-term and long-term incidents and modelled them separately [22]. An important thing to note is that in all of the previous studies using US traffic accident dataset, accident severity and textual description of the accident has been used as a feature. These features are subjective and may not be available in real life scenarios.

Reducing traffic accidents and its impact on traffic movement is a top public safety priority. However, the majority of research on traffic accident analysis and prediction has relied on small-scale or filtered datasets with limited coverage, which diminishes their usefulness and impact. In addition, the available large-scale datasets are either private, out-of-date, or lack crucial contextual information such as environmental stimuli (weather, landmarks, etc). This problem is more prevalent in Bangladesh, where there are no comprehensive and enhanced open access datasets. In this sense, a comprehensive database containing US accident records and comparable weather and temperature patterns can be utilised. The dataset contains 2.25 million incidents of crashes involving vehicles [20]. As the environment and climate of Texas are most comparable to those of Bangladesh, we chose Texas data for this investigation.

The present study intentionally omits the utilization of accident descriptions and categories in the prediction of traffic accident duration. This decision is substantiated by the inherent complexity and often unpredictable nature of traffic accident situations, which make the on-the-spot categorization of accident severity and the collection of objective descriptions challenging, if not unrealistic. Moreover, the reliance on such subjective and often inconsistent information could potentially compromise the integrity and accuracy of the prediction model. Therefore, we argue that these elements should not form an integral part of a comprehensive and effective response strategy.

We explore the utilization of various machine learning (ML) methodologies with the objective of predicting the duration of traffic accidents. Our methodological framework adopts a bi-modal strategy, categorizing accidents into either short-term or long-term, based on the average accident duration. This allows us to tackle the problem from both a classification and regression standpoint.

Our application of multiple ML algorithms, including Random Forest, XGBoost, and CatBoost, was specifically aimed at predicting the binary duration of traffic accidents, distinguishing them into short-term and long-term categories. Furthermore, within these categories, we leveraged CatBoost, RFCNN (Random Forest Convolutional Neural Network), and LightGBM (Light Gradient Boosting Machine) models to predict the actual duration of the accidents.

An evaluation of the models' performances facilitated the selection of the most effective algorithms. From this analysis, we were able to devise a novel, integrated methodology that closely emulates the real-world circumstances within an intelligent transport system.

In this approach, the combined test set is initially used to predict if an accident falls into the long-term or short-term category. Following this classification, the associated features are inputted into separately trained models to predict the expected traffic delay that would be caused by the accident.

The comprehensive performance of this end-to-end approach was then evaluated, with the Root Mean Square Error (RMSE) and Mean Absolute Error (MAE) serving as the key metrics. The results indicate that our novel, integrated approach provides a promising avenue for predicting traffic accident durations and subsequent delays, contributing to more effective traffic management within intelligent transport systems.

The primary contributions offered by this manuscript encompass several significant aspects:

1. A comprehensive benchmarking process has been executed, encompassing multiple well-established machine learning models. This thorough approach has facilitated the identification of the model yielding the highest performance and optimally tuned parameters for the specific problem under scrutiny.
2. The employment of weather parameters, along with road condition and location, as exclusive features, serves a dual-purpose advantage. Primarily, it paves the way for an implementable solution in predicting the impact of an accident on the road traffic network, based solely on the accident location. This, in turn, equips the road traffic authority with the means to respond more effectively. Furthermore, no prior studies have exclusively leveraged weather conditions and road parameters for this purpose, thereby offering a novel perspective on the importance of these static parameters in predicting traffic delays due to accidents.
3. A novel Bi-modal methodology, representing an end-to-end pipeline, has been introduced. This innovative approach integrates the best-performing classification model with the most effective regression model, thereby producing a combined output. This methodology fosters enhanced accuracy in predicting traffic accident duration and subsequent delays, offering a holistic view of the problem.

These advancements not only contribute to the existing body of knowledge but also pave the way for more effective traffic management strategies within intelligent transport systems.

**Data Description**

**Data Description Table**

The dataset employed in this study is a comprehensive collection of traffic accident records that span across 49 states within the United States. The process of data compilation has been ongoing since February 2016 and draws upon several data sources. These sources include a variety of APIs offering real-time traffic event data, as well as state and national departments of transportation, law enforcement agencies, traffic sensors situated within road networks, and traffic cameras.

These varied and extensive sources have enabled the creation of a rich, multifaceted dataset, which, as of now, encompasses approximately 1.5 million individual accident records [23]. The dataset can be accessed via the following link: US-Accidents: A Countrywide Traffic Accident Dataset - Sobhan Moosavi (smoosavi.org).

For this task we are using data from February 2016 to December 2021 collected MapQuest Realtime Traffic Data Collector [24] which is mentioned as source 1 in the dataset for the state of Texas. The detailed description of the columns of the data are provided in Table:1.

Table 1: Data Description

| # | Attribute | Description | Nullable |
|---|---|---|---|
| 1 | ID | This is the accident record's unique identifier. | No |
| 2 | Severity | The severity of the accident is indicated by a number between 1 and 4, with 1 indicating the least impact on traffic (i.e., a short delay as a consequence of the accident) and 4 indicating a significant impact on traffic (i.e., a long delay). | No |
| 3 | Start_Time | Displays the accident's commencement time in the local time zone. | No |
| 4 | End_Time | Displays the accident's end time in the local time zone. End time refers to the point at which the accident's effect on traffic flow was eliminated. | No |
| 5 | Start_Lat | Displays the GPS latitude of the starting point. | No |
| 6 | Start_Lng | Displays the GPS longitude of the starting point. | No |
| 7 | End_Lat | Displays the latitude in GPS coordinates of the destination. | Yes |
| 8 | End_Lng | Displays the longitude of the endpoint as a GPS coordinate. | Yes |
| 9 | Distance(mi) | The length of the roadway impacted by the collision. | No |
| 10 | Description | The extent of the roadway that was affected by the accident. | No |
| 11 | Number | Displays the street number in the address. | Yes |
| 12 | Street | Street name is displayed in the address field. | Yes |
| 13 | Side | Displays the relative street side (Right/Left) in the address field. | Yes |
| 14 | City | Displays city in the address field. | Yes |
| 15 | County | Displays county in the address field. | Yes |
| 16 | State | State is displayed in the address field. | Yes |
| 17 | Zipcode | Displays the postal code in the address field. | Yes |
| 18 | Country | Displays the nation in the address field. | Yes |
| 19 | Timezone | Displays the timezone based on the accident's location (eastern, central, etc.). | Yes |
| 20 | Airport_Code | Indicates the airport-based weather station that is closest to the accident site. | Yes |

| 21 | Weather_Timestamp | Displays the timestamp (in local time) of the weather observation record | Yes |
|---|---|---|---|
| 22 | Temperature(F) | Displays the current temperature in Fahrenheit | Yes |
| 23 | Wind_Chill(F) | Displays the wind chill in degrees Fahrenheit. | Yes |
| 24 | Humidity(%) | Displays the relative humidity in percent | Yes |
| 25 | Pressure(in) | Displays air pressure in centimetres. | Yes |
| 26 | Visibility(mi) | Displays the visibility in miles. | Yes |
| 27 | Wind_Direction | Indicates airflow direction. | Yes |
| 28 | Wind_Speed(mph) | Displays wind speed (in miles per hour). | Yes |
| 29 | Precipitation(in) | Displays the precipitation quantity in inches, if any exists. | Yes |
| 30 | Weather_Condition | Displays the current weather (rain, snow, thunderstorm, mists, etc.). | Yes |
| 31 | Amenity | . A POI annotation that denotes the presence of a nearby amenity. | No |
| 32 | Bump | A POI annotation indicating the presence of a speed bump or hump in the vicinity. | No |

| # | Name | Description | |
|---|---|---|---|
| 33 | Crossing | A POI annotation indicating the existence of a crossing in the vicinity. | No |
| 34 | Give_Way | A POI annotation indicating the presence of give_way in the area. | No |
| 35 | Junction | A POI annotation indicating the existence of a nearby intersection. | No |
| 36 | No_Exit | A POI annotation indicating the presence of no_exit in the area. | No |
| 37 | Railway | A POI annotation indicating the presence of a railway in the vicinity. | No |
| 38 | Roundabout | A POI annotation indicating the presence of a roundabout in the vicinity. | No |
| 39 | Station | A POI annotation that denotes the presence of a station in the vicinity. | No |
| 40 | Stop | A POI annotation indicating the presence of a halt in the vicinity. | No |
| 41 | Traffic_Calming | A POI annotation indicating the presence of traffic_calming in the area. | No |
| 42 | Traffic_Signal | A POI annotation indicating the presence of a traffic_signal in the vicinity. | No |

| 43 | Turning_Loop | A POI annotation that denotes the nearby presence of turning_loop. | No |
| 44 | Sunrise_Sunset | Displays the time of day (day or night) based on sunrise and sunset. | Yes |
| 45 | Civil_Twilight | Displays the time of day (day or night) according to civil twilight. | Yes |
| 46 | Nautical_Twilight | Displays the time of day (day or night) according to nautical twilight. | Yes |
| 47 | Astronomical_Twilight | Displays the time of day (dawn or dusk) based on astronomical twilight. | Yes |

**Feature Selection**

The impact of the model and the training time may be diminished by any duplicate or irrelevant features. As a result, feature selection might potentially improve performance by increasing interpretability, hastening model training, and reducing noise and overfitting in addition to reducing noise and overfitting.

When a traffic accident is confirmed, the main goal of this research is to investigate how to more accurately anticipate how long its effect will last on the road by using a few readily accessible variables. Therefore, factors like the accident description and accident severity that can only be learned after the fact will be disregarded. The screening process resulted in the selection of the time attribute, the points of interest (POI) attribute, the weather attribute, the location attribute, the traffic attribute and others for this article.

Finally, a total of 27 variables were taken as input, these are: 'Distance(mi)', 'Temperature(F)', 'Wind_Chill(F)', 'Humidity(%)', 'Pressure(in)', 'Visibility(mi)', 'Wind_Direction','Wind_Speed(mph)', 'Precipitation(in)', 'Weather_Condition', 'Amenity','Bump', 'Crossing', 'Give_Way', 'Junction', 'No_Exit', 'Railway','Roundabout', 'Station', 'Stop', 'Traffic_Calming', 'Traffic_Signal','Sunrise_Sunset', 'Civil_Twilight', 'Nautical_Twilight', 'Astronomical_Twilight'.

**Data Processing**

**Dealing with Missing Data**

Data loss is unavoidable because it can occur during data recording and transfer as well. The mean of the numerical attribute was used to fill in missing values such as temperature, wind_chill, humidity, pressure, and visibility. For categorical features such as wind_direction, sunrise_sunset, and weather_condition, the feature was first turned into a binary variable, and then the missing values were filled using the feature's mode value.

**Feature Calculation**

The main feature calculation performed was for calculating accident duration in minutes,
The formula used here was shown in Equation 1:

$$\text{Time Difference (in minutes)} = \frac{(End\ time - Start\ time)}{60} \quad [1]$$

After calculating the minutes, we performed statistical analysis to see how the time duration was distributed. First, we check the maximum, minimum, average and standard deviation value.

Table 2: Time difference calculation

| Parameter (in mins) | In minutes | After removing outlier |
|---|---|---|
| Maximum | 961379.0 | 360.0 |
| Minimum | 3.0 | 27.51 |
| Average | 202.689 | 164.46 |
| Standard Deviation | 3159.15 | 120.64 |

We can see from the initial analysis that there are plenty of outliers which makes the average and standard deviation go haywire. To resolve this issue we drop the 5% data on both ends of the spectrum, ie we only keep the middle 90% of the data. After removing these outliers the data distribution becomes quite regular as can be seen in Table 2 and Figure 1.

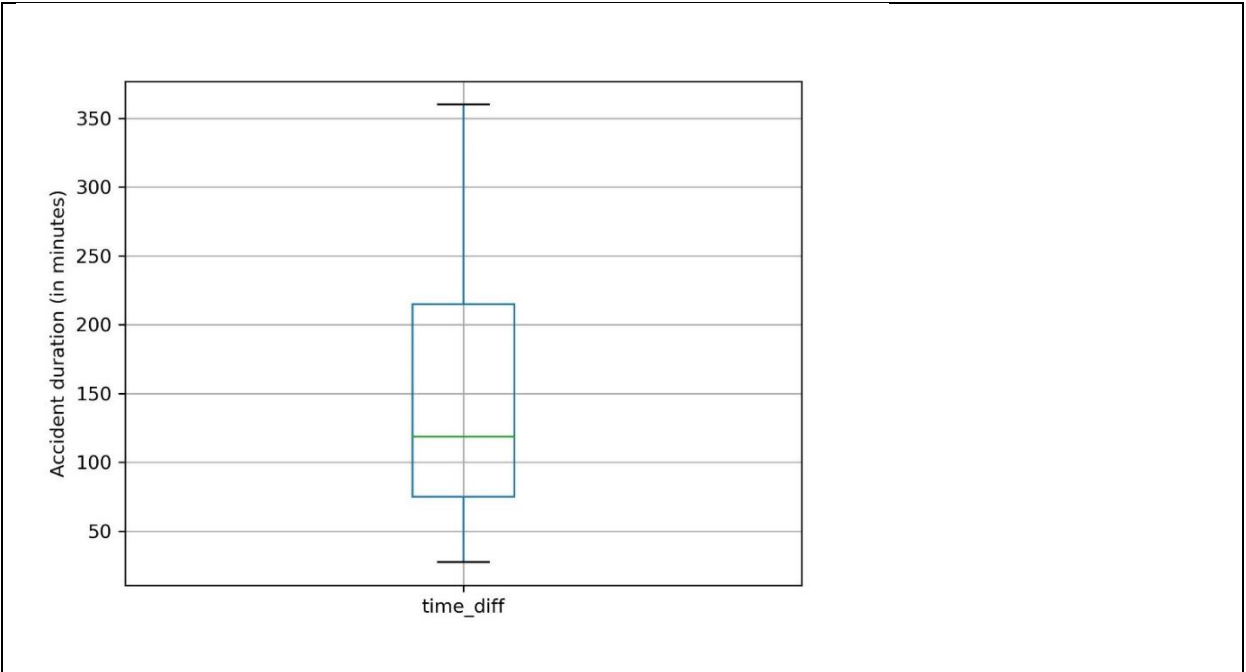

Figure 1: Boxplot of accident duration

In Figure 1, the green line represented the median of the feature after removing the outliers. The mean of the distribution is between 100 and 150, while the first quartile is at 75 and 3$^{rd}$ quartile is at 210. The total of the data is between 20 to 360.

From this plot we applied a conditional calculation, to divide the accident durations with less than 164 minutes as short duration accident (labelled 1) and greater than 164 minutes as long duration events (labelled 0). Similar sort of procedure was also taken in a previous work[25].

## Correlation Matrix

As the dataset was huge, the correlation matrix only shows small correlation between the variables. Correlation matrix targeting accident duration and inputs has been produced.

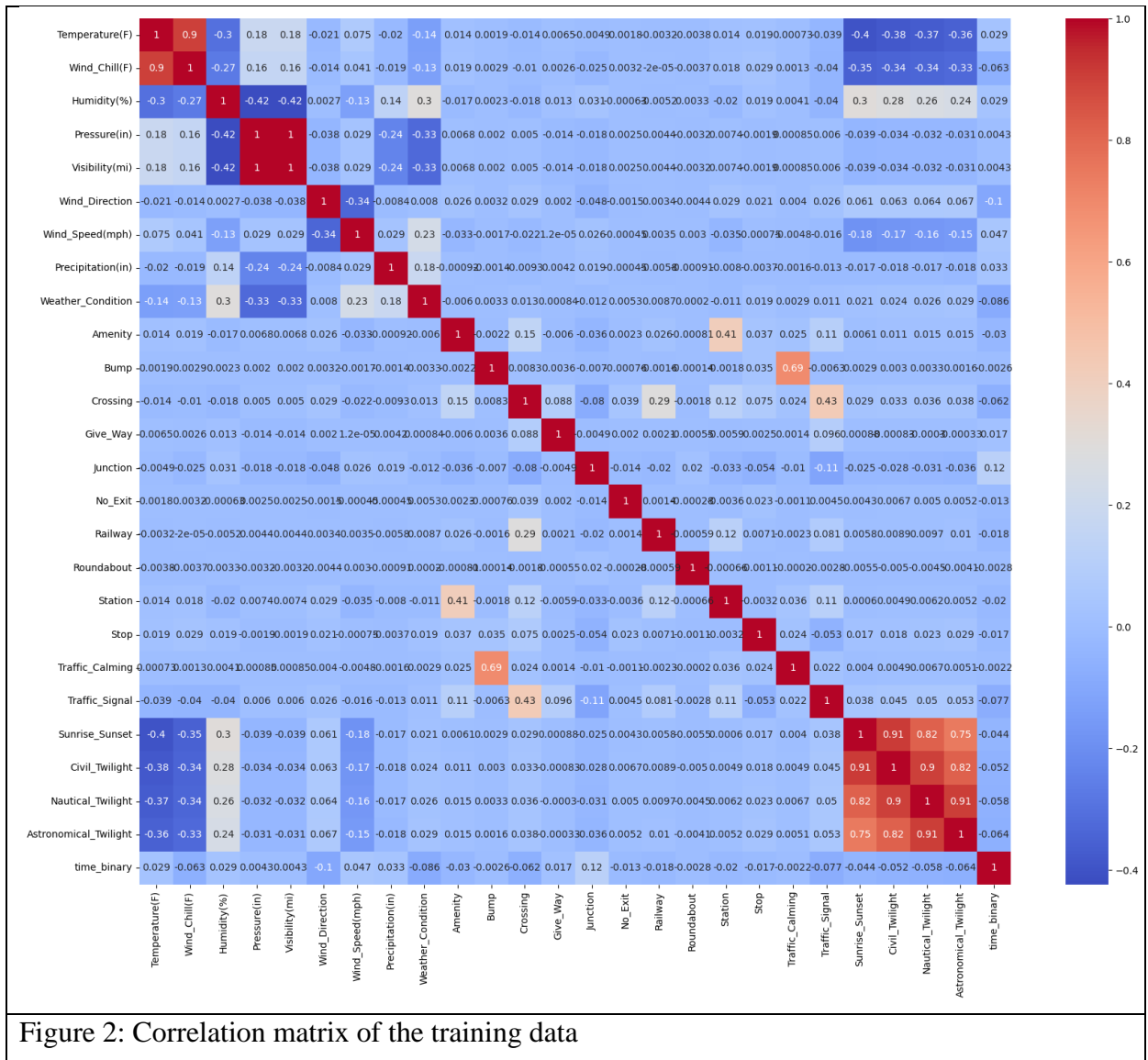

Figure 2: Correlation matrix of the training data

Figure 2 shows the accident duration correlation matrix with respect to the inputs; there is a modest link between the output response accident duration and wind direction, weather condition, junction, and traffic signal. Wind direction, weather conditions, and traffic signals all have a negative effect on output responsiveness, whereas junctions have a positive effect. Based on this matrix, we may deduce that the four input attributes have the greatest influence on the output response.

**Data Distribution**

Based on the correlation matrix it has been found out three numerical and two categorical feature has highest impact on the output response accident duration. The first numerical feature is Temperature.

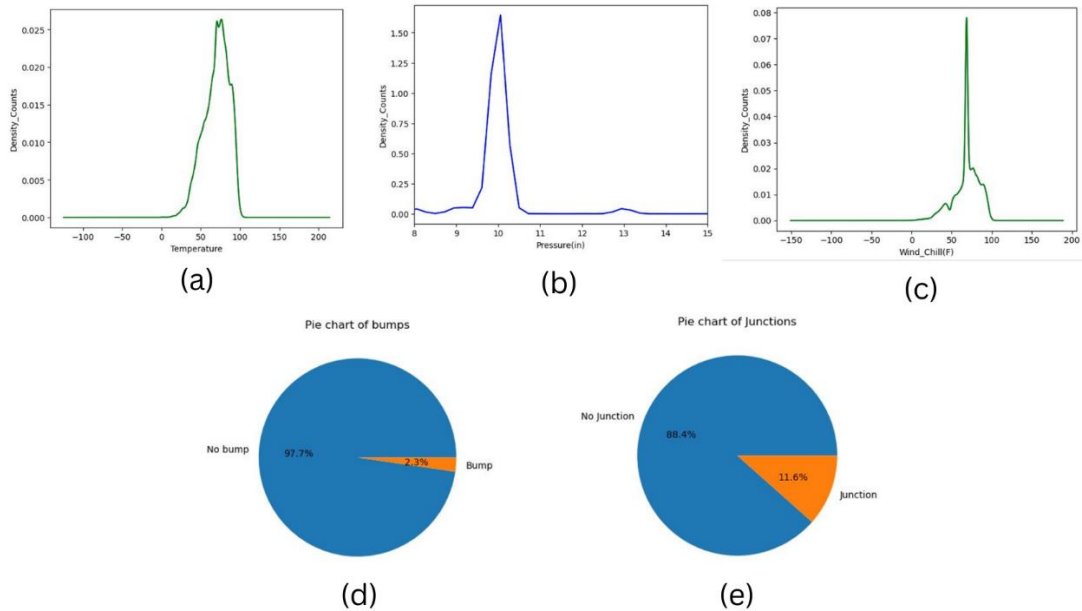

Figure 3: Data distribution plot of the most significant inputs (a) Temperature (b) Pressure (c) Wind chill, (d) Bumps and (e) Junctions

Based on the density plot of temperature in Figure 3(a), we have plotted temperature on the X-axis and density were plotted in the Y-axis. Temperature density was high between 50-80 temperature and temperature has reasonably high impact on the output responses.

Similarly, the other two numerical feature pressure and wind-chill are shown using density plot in Figure 3(b), 3(c).

For the categorical data, using the Pie chart plot we plotted the data based on two categories. For the 'bump' feature the category was first divided into two and then the pie chart was plotted. Similarly, the ither feature 'junction' was also plotted using pie chart in Figure 3(d),3(e).

**Analytical Result**

**Evaluation Index**

Three groups of evaluation indicators, mean absolute error (MAE) and root mean square error (RMSE) has been used to evaluate the results.

The representing equations Equation. (2) - (3) of these parameters are given below

$$\text{RMSE} = \sqrt{\frac{\sum_1^n (y-x)^2}{n}} \quad [2]$$

$$\text{MAE} = \frac{1}{N} \sum_{i=1}^{n} \frac{|Actual - predicted|}{Actual} \quad [3]$$

For classification tasks, popular evaluation metrics like, Accuracy, precision, recall and F1-score have been used, the parameter definitions are given below.

$$Accuracy = \frac{TP + TN}{TP + FP + TN + FN}$$

$$Precision = \frac{TP}{TP + FP}$$

$$Recall = \frac{TP}{TP + FN}$$

$$F1score = \frac{2 * TP}{2 * TP + FP + FN}$$

**Model Implementation (Data Partitioning (train-test-validation), Model Diagnostics)**

**Data Partitioning**

The filtered dataset after outlier removal had 134629 accident data points. Within this dataset we selected 75% as training and 25% as validation. Using this data, we will use a bimodal approach to predict accident duration, first we will use a classification model to predict whether the accident duration was long and short, and then use two different models for predicting the exact time amount of the accident.

**Classification model selection:**

We used the Random Forest, XGBoost, and CatBoost models for classification. The models were chosen due to their suitability for predicting on tabular data. The performance of these models are shown in Figure 4 and Table 3.On the test dataset, the RF model achieved a remarkable 83% classification accuracy. This demonstrates the model's proficiency in classifying instances precisely. The precision and recall scores further demonstrate the model's ability to identify positive and negative instances with equal accuracy. These metrics indicate the model's ability to reduce both false positives and false negatives, a crucial quality for classification tasks with significant consequences.

A comprehensive set of performance evaluation metrics was used to evaluate the efficacy of the XGBoost classification model. These metrics comprise various aspects of the model's predictive capabilities, allowing for a nuanced comprehension of its performance across multiple dimensions. Accuracy, precision, recall, and F1-score are utilized as evaluation metrics. The performance results of the XGBoost classification model demonstrate its exceptional ability to resolve the inherent complexities of the dataset. The model's 81% accuracy indicates its capacity to correctly classify instances. The precision score of 82% demonstrates the model's ability to minimize false positive predictions, whereas the recall score of 81% demonstrates its skill in identifying genuine positive instances. The F1-score, which combines precision and recall, was calculated to be 80%, illustrating the model's balanced performance in attaining both precision and recall goals.

The CatBoost classifier also demonstrated a remarkable 83% overall accuracy on the test dataset. This demonstrates the model's capability to classify instances precisely into their respective classes. The precision and recall scores further validate the model's ability to identify both positive and negative instances in a balanced manner. These metrics indicate the model's ability to minimize both false positives and false negatives, which is crucial for classification tasks with high misclassification costs.

The RF classifier outperformed the other two models based on the confusion matrix and accuracy, precision, recall, and F1-score.

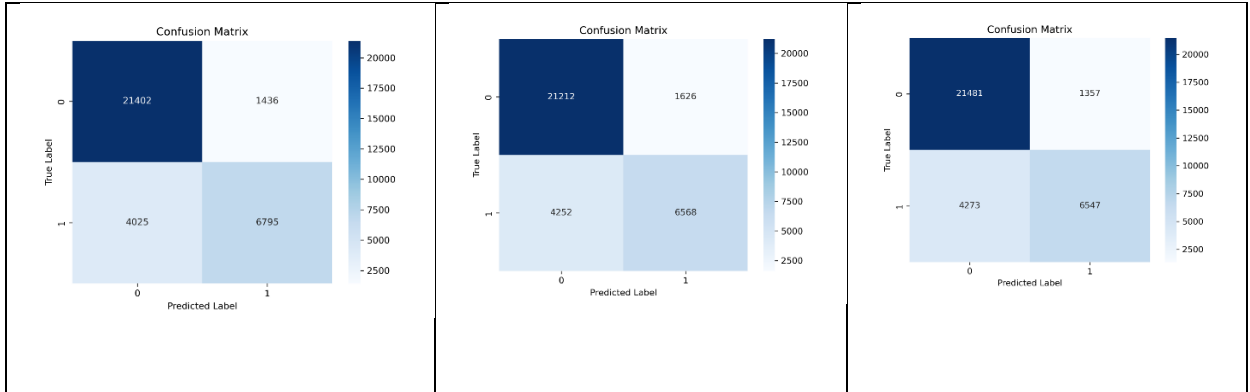

Figure 4: Confusion Matrix of RF, XGBoost and CatBoost consecutively

Table 3: Binary classification results of accident duration prediction

| Model | Accuracy | precision | recall | f1-score |
|---|---|---|---|---|
| **Random Forest** | **0.83** | **0.84** | **0.83** | **0.83** |
| XGBoost | 0.81 | 0.82 | 0.81 | 0.80 |
| CatBoost | 0.83 | 0.83 | 0.83 | 0.82 |

**Regression model selection:**

RFCNN, CatBoost, Light Gradient Boosting Machine, and LSTM models were utilized for the Regression task. The selection of these models was based on their superior performance in regression tasks. The structure of the LSTM model is shown in Figure 5.

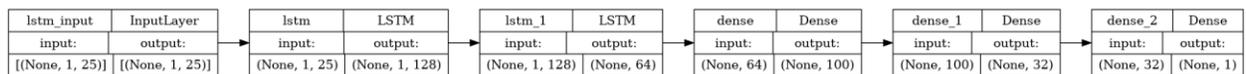

Figure 5: LSTM model structure

To evaluate the efficacy of the regression algorithm, we employed standard evaluation metrics such as root mean squared error and mean absolute error. These metrics provide insight into various aspects of regression performance, including the precision of predictions and the model's goodness-of-fit. The performance of these models are shown in Table 4. LightGBM regression consistently demonstrated competitive performance in terms of accuracy across all datasets used in our experiments. It achieved low RMSE (32.9128 for the short term and 28.9414 for the long term) and MAE values (26.15 for the short term and 13.69 for the long term), indicating its ability to make precise predictions with minimal deviation from the actual values. This suggests that LightGBM regression effectively models the relationship between the input features and the target variable and provides a reasonable fit to the underlying regression problem.

In terms of accuracy, RFCNN, CatBoost, and LSTM also demonstrated competitive performance. As shown in Table 4, they obtained low RMSE and MAE values, indicating their ability to make accurate predictions with minimal deviation from the actual values. This indicates that these algorithms capture the underlying patterns and relationships in the data effectively, resulting in accurate predictions.

Table 4: Accident duration prediction performance regression results

| Model | Short Term | | **Long Term** | |
|---|---|---|---|---|
| LightGBM | RMSE | **32.9128** | RMSE | **28.9414** |
| | MAE | **26.15** | MAE | **13.69** |
| RFCNN | RMSE | 34.9 | RMSE | 34.22 |
| | MAE | 27.5 | MAE | 17.60 |
| CatBoost | RMSE | 33.79 | RMSE | 32.93 |
| | MAE | 27.51 | MAE | 16.59 |
| LSTM | RMSE | 34.97 | RMSE | 33.59 |
| | MAE | 28.06 | MAE | 17.94 |

According to the regression analysis, the LightGBM model performed better than the other three models.

Therefore, LightGBM model achieved the best predictive performance on the regression assignment for both short- and long-term predictions; examples of its predictions on the test data are provided.

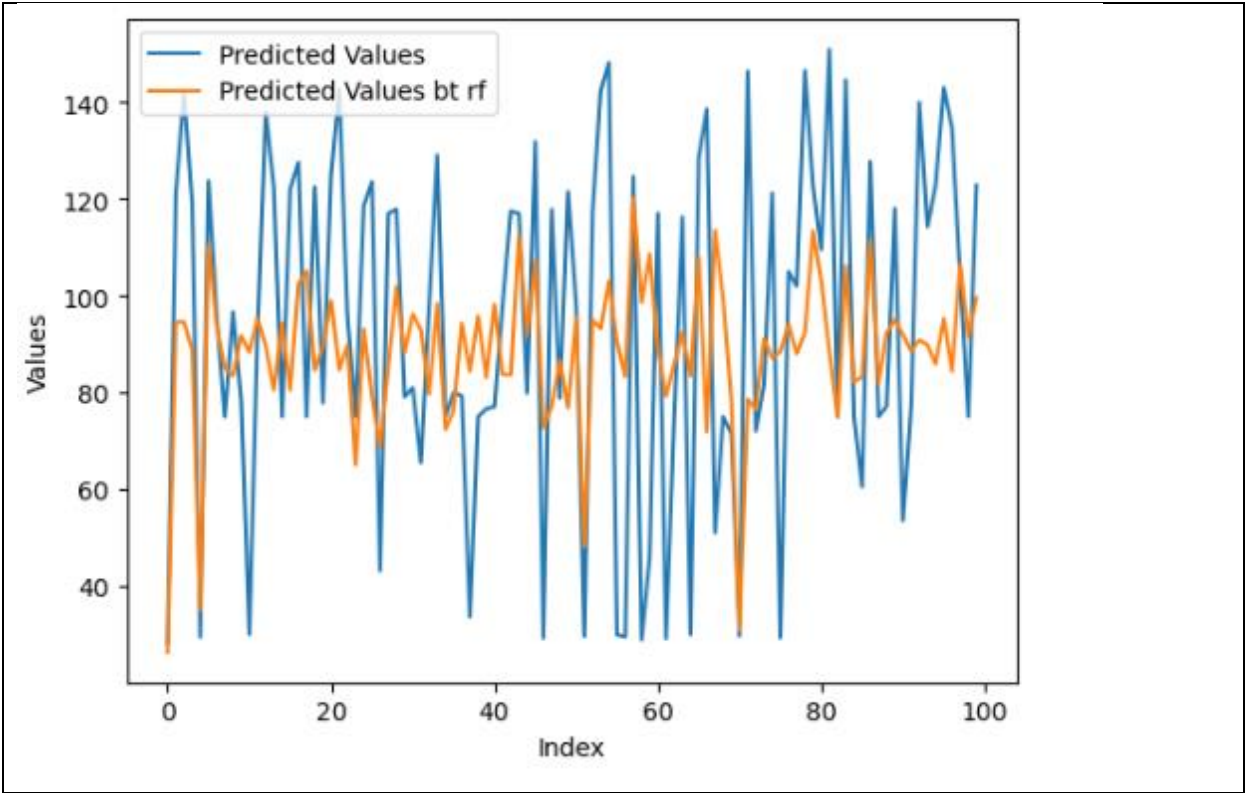

Figure 6: Short duration accident time prediction using LightGBM model for first 100 data

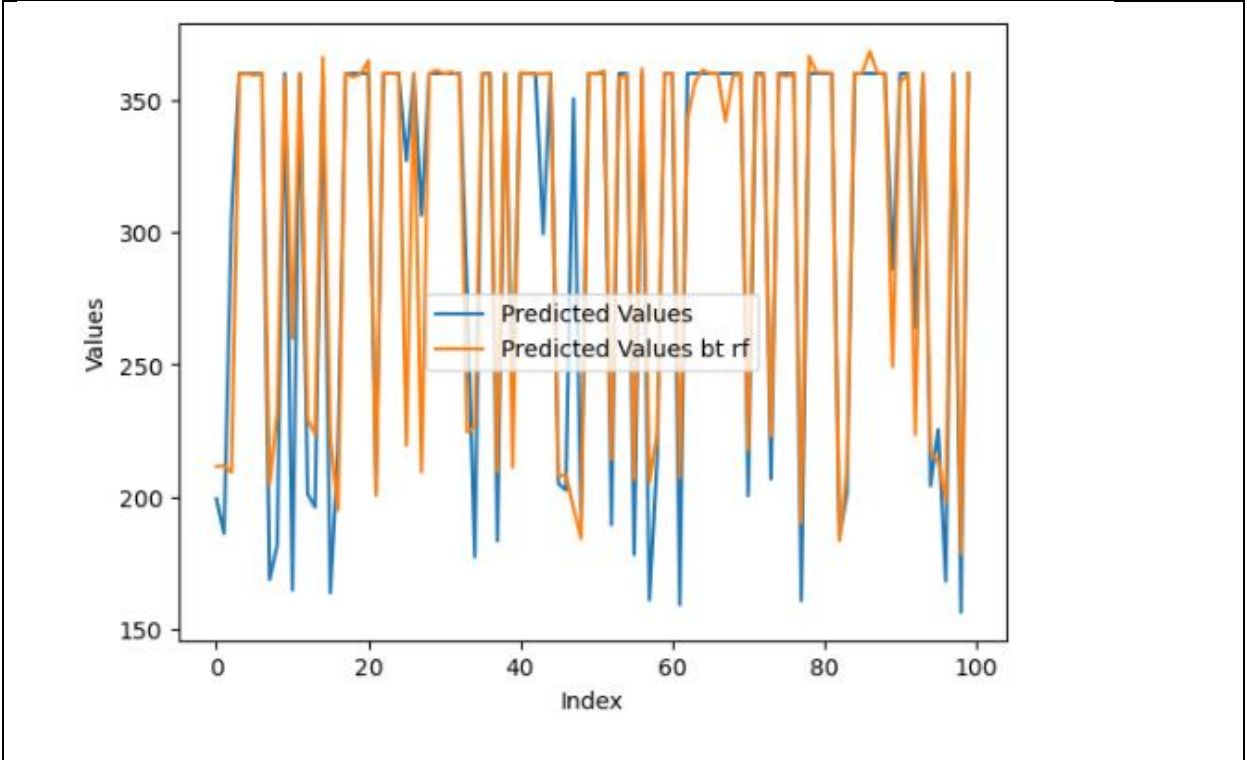

Figure 7: Long duration accident time prediction using LightGBM model for first 100 data

Upon meticulous examination of Figure 6 and Figure 7, it is evident that the level of correlation between the two sets of values is remarkably high. This observation demonstrated the model's exceptional predictive capabilities and its ability to capture the data's fundamental patterns and trends.

Initially, it is essential to observe that the predicted values closely match the actual data points across the entire graph's range. Consistently reflecting the actual values, the predictions exhibited a high degree of accuracy throughout. This consistency demonstrated the robustness of the model in capturing the intricate relationships and dependencies inherent to the dataset.

In addition, the graph revealed an exceptional degree of regularity in the predicted values. The predicted curve exhibited a continuous and coherent trajectory, similar to the actual data's pattern. This smoothness indicated that the model has effectively encapsulated the underlying dynamics of the dataset, including any temporal or spatial dependencies, and has utilized this information to generate predictions that progress consistently and realistically over time.

**Comparison with other works**

As mentioned earlier, traffic accident duration prediction on the dataset used here has been performed using different approaches. In Table 5, the results of the previous approaches will be compared with the results achieved in this work.

Table 5: Comparison with existing results

| Model | Dataset | RMSE (in minutes) | MAE (in minutes) | Reference |
|---|---|---|---|---|
| N-BEATS | All states, including severity data | Combined: 120 | Combined: 9.6 | [26] |
| GBDT with textual encoding | San Francisco, California State, dataset enhanced with data from Vehicle Detection Stations (VDS) | Combined: 41.89 | Combined: 65.03 | [21] |

| Model | Dataset | RMSE | MAE | Ref |
|---|---|---|---|---|
| Heterogenous Ensemble | Full dataset, using 98% data for training | Combined: 65.2087 | Combined: 30.74 | [7]. |
| IEO-ML pipeline | San-Francisco road network | Short Term: 9.34  Long Term: 23.69 | | [22]. |
| LGBM | Texas State | Short Term: **32.91**  Long Term: **28.94** | Short Term: **26.15**  Long Term: **13.69** | Proposed |

Existing strategies for traffic accident duration prediction can largely be bifurcated into two distinctive parts. Primarily, the prediction of duration has been carried out by aggregating all accidents, a strategy that, despite being convenient to implement, detrimentally impacts the accuracy of predictions. This is manifest in the N-BEATS approach, with Root Mean Square Error (RMSE) values reaching up to 120 minutes [26]. An alternative approach achieved an RMSE value of 41.89, albeit on a limited subset of the California dataset, supplemented by Vehicle Detection System (VDS) data [21]. A more recent study utilized a heterogeneous ensemble model to attain an RMSE score of 65.21 on the complete dataset, but this method involved utilizing 98% of the data for training and merely 1% for testing (the latest month of data). While the accuracy was ostensibly enhanced in comparison to previous models, the limited test data render this approach deficient in certain aspects [7].

The most exhaustive task thus far was undertaken by Grigoriev et al., who leveraged data from the road networks of San Francisco, Sydney, and Victoria to evaluate a combination of bi-level, combined, and cross-training approaches using pipeline and fusion models [22]. For the purposes of comparison, the outcomes of the bi-modal model have been incorporated in the comparison

Table 5. The impressive results obtained on the San Francisco Road network can primarily be attributed to the limited geographical extent of the data collection area. As aforementioned, all the approaches evaluated herein utilize accident severity and, in the N-BEATS and GBDT models with textual encoding, they employ textual description data, which has been deliberately omitted in our proposed model to enhance usability and applicability.

Our proposed model, which utilizes Light Gradient Boosting Machine (LightGBM), exhibits results that are not only comparable but, in certain instances, superior to previous models, and this is achieved without the use of accident severity and accident description data and applied to the entirety of Texas. The RMSE for short-term predictions was recorded as 32, while the RMSE for long-term predictions was further reduced to 28.

**Combined Pipeline: Structure and Comparison**

The development of an integrated end-to-end pipeline employing both random forest classification and LightGBM regression models with only static variables like weather and road conditions without any accident context data has proved to be a major advancement in accident duration prediction. This pipeline consists of numerous intricate stages that improve the accuracy and efficacy of predictions.

The initial dataset is subjected to an exhaustive analysis, taking into account numerous accident-related factors and attributes. This data set was then separated into two categories: long-term and short-term accidents. This partitioning strategy enabled a more focused and individualized approach to prediction for each specific duration.

The first component of the pipeline was a classification model based on random forest. This potent algorithm classified accident instances into long-term and short-term categories using a collection of decision trees. This classification task was ideally suited to the random forest's capacity to manage complex relationships and capture feature interactions. Utilizing a large number of decision trees and their collective predictions, the model accurately categorizes accidents according to their duration categories.

As soon as the classification phase concludes, the pipeline transitions without interruption to the regression phase. Here, the LightGBM regression model was chosen due to its demonstrated superiority in handling large datasets and identifying intricate data patterns. LightGBM accurately predicted the duration of long-term and short-term accidents independently by utilizing gradient enhancing techniques.

The complexity of this integrated pipeline lay not only in its use of sophisticated machine learning models, but also in its meticulous coordination of multiple stages and considerations. The selection of appropriate models for classification and regression tasks, the strategy for data partitioning, and the seamless integration of the models into a unified pipeline all contributed to the complexity and refinement of this approach.

To contextualize the efficacy of our proposed pipeline, a comparison was made with a prior research work that employed a bimodal approach as shown in Table 6. The said work proposed both a pipeline and a fusion model, with the fusion approach yielding lower Root Mean Square Error (RMSE) values. However, it is important to highlight that this approach is primarily based on data derived from a relatively smaller segment of California State, specifically the San Francisco Road networks. Additionally, this approach incorporated accident context information in the prediction process.

Moreover, the fusion approach requires the concurrent utilization of four distinct models, rendering it an over-engineered solution. While this might lead to an enhanced level of accuracy, its application in real-world scenarios can potentially pose challenges.

Our model, which is predicated solely on static features, achieves results that are comparable to those derived from the analysis of a more diminutive and specific road network dataset, as used in the aforementioned study. This is even more noteworthy given that our model was applied to the entirety of a state. Thus, our approach not only offers promising performance but also demonstrates practical applicability at a broader scale.

Table 6: Combine result of accident duration prediction.

| Proposed Pipeline | Texas State data | RMSE | 94.70 minutes |
|---|---|---|---|
|  |  | MAE | 68.6785 minutes |
| Previous pipeline | SF road network | RMSE | 78 minutes |
| Previous fusion |  | RMSE | 59 minutes |

**Prescriptive Insight**

In an effort to decipher the complex relationship between input features and output predictions, we employed the potent techniques of tree visualization and SHAP value analysis. The complex and multifaceted character of the binary accident duration prediction model can be grasped through tree visualization, a visual representation of the decision-making process [shown in Figure 8] . Due to the incredible complexity and richness of the tree, it is impossible to visualize it in its entirety, leaving us with only glimpses of its intricate branches and nodes.

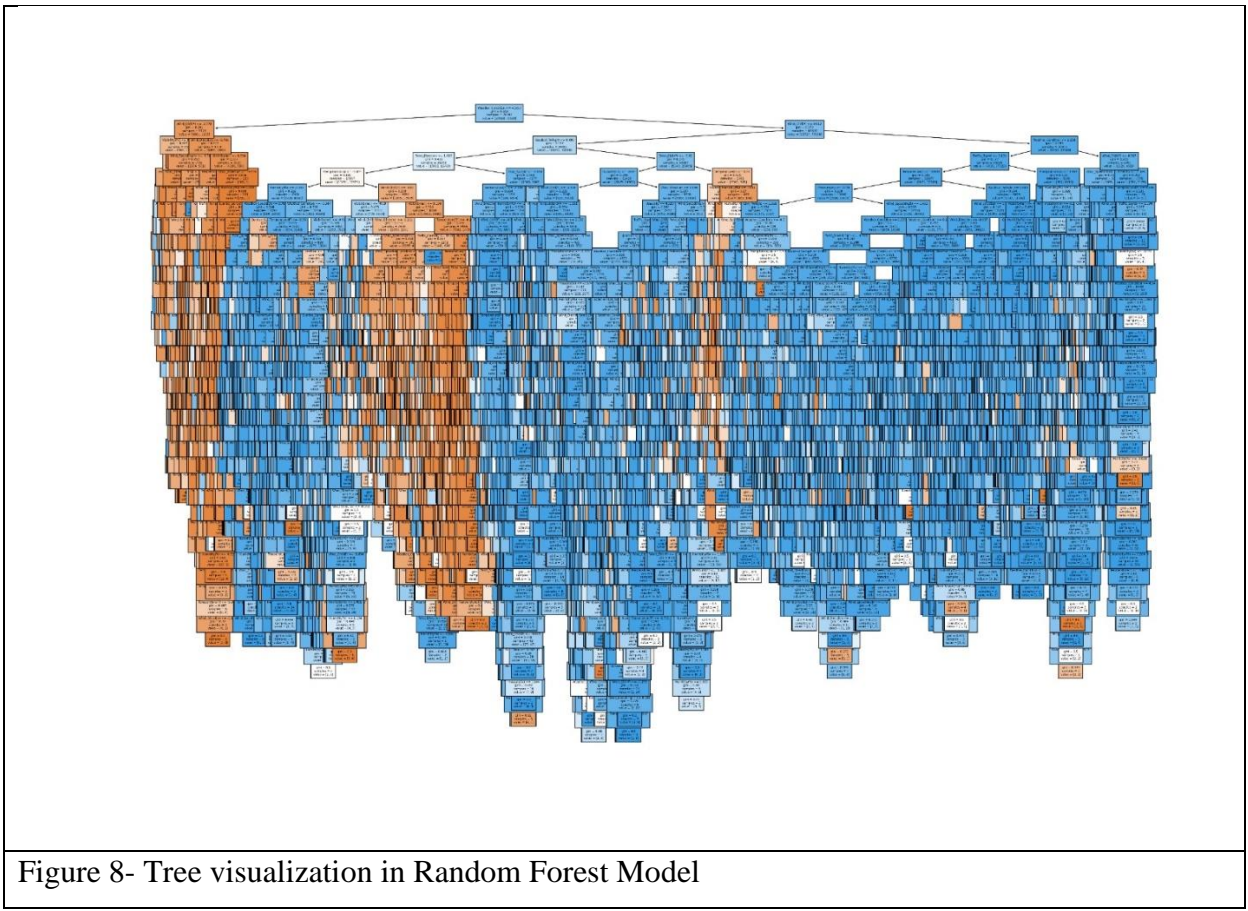

Figure 8- Tree visualization in Random Forest Model

Using SHAP (Shapley Additive Explanations) values, we performed a comprehensive analysis to obtain a deeper understanding of the factors influencing accident duration prediction. SHAP values provide a robust framework for interpreting and comprehending the contribution of each feature to the final predictions generated by our selected models, random forest and LightGBM.

From the two figures [Figure 9 and 10], it is evident that the feature 'wind_chill(F)' and 'precipitation(in)' and 'Wind_Direstion' exhibited the highest average absolute SHAP value, indicating its significant influence on both long and short term accident duration predictions.

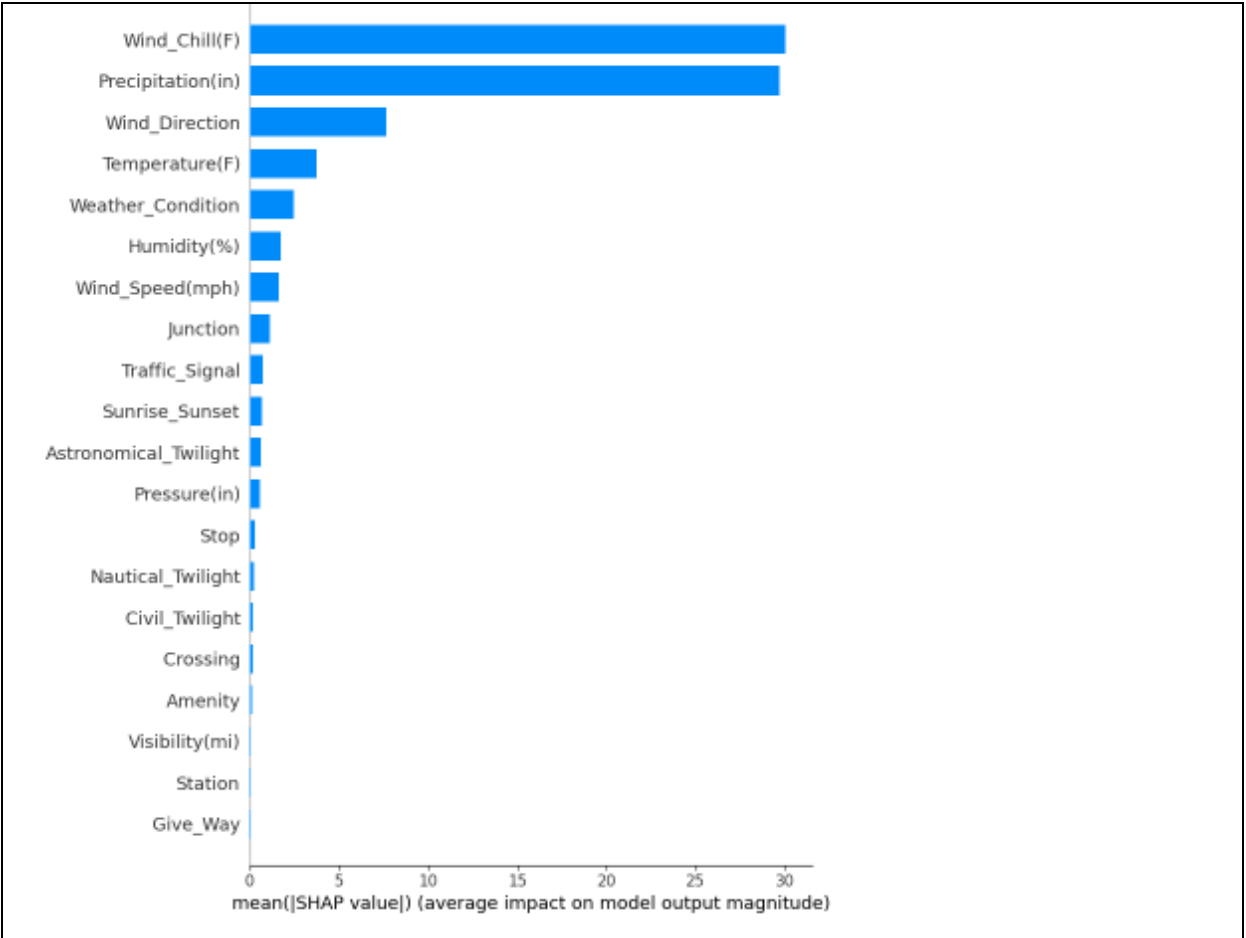

Figure 9: SHAP values for input features in short period accident duration prediction

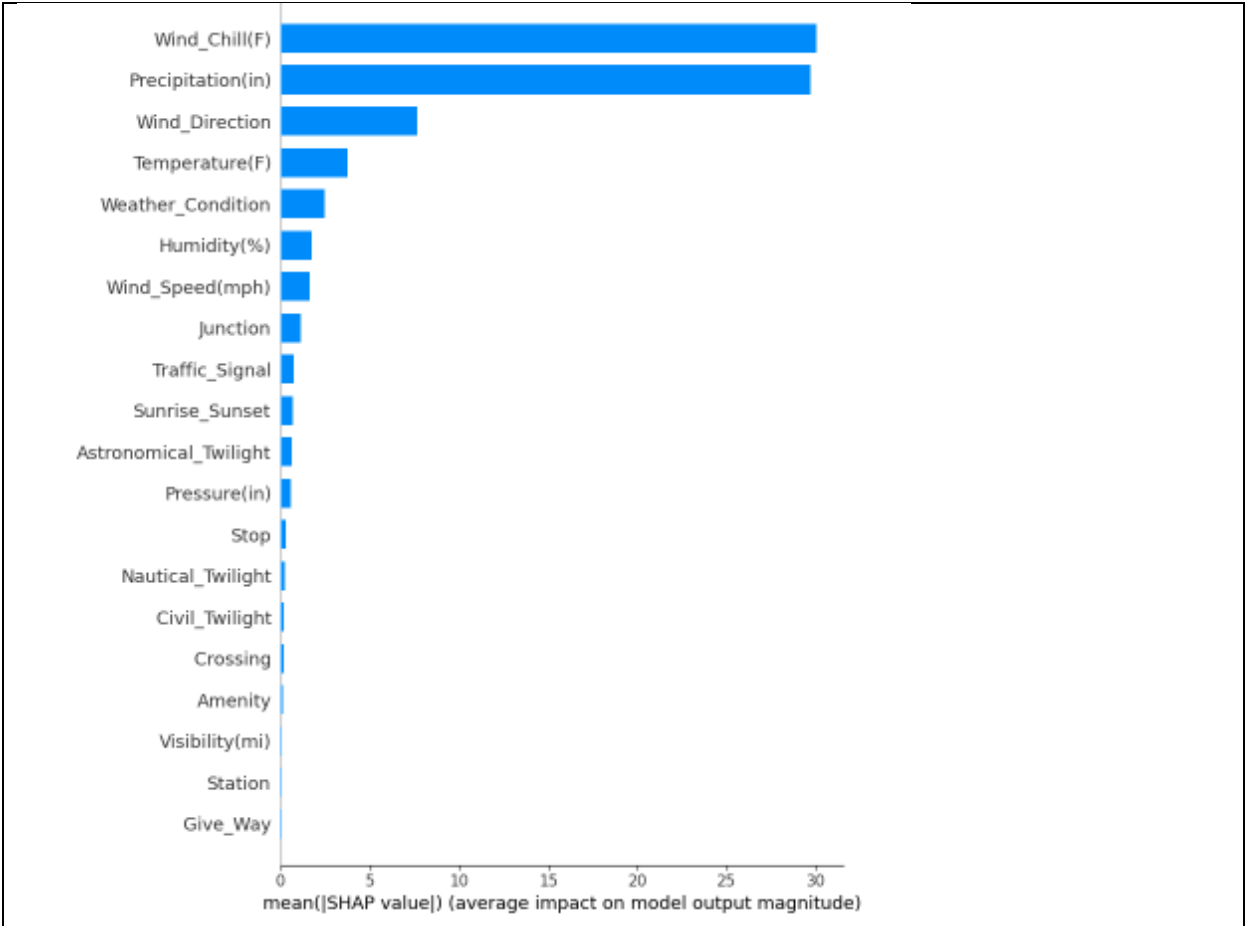

Figure 10: SHAP values for input features in long period accident duration prediction

Overall, the SHAP value analysis provided insightful information regarding the relative relevance of features in predicting accident duration. By utilising SHAP values, we improved the interpretability and explanability of our models by gaining a deeper understanding of how these characteristics affect accident duration predictions.

**Conclusion**

This study is centered on the examination of traffic accident durations in the state of Texas, specifically during the initial stages of the incidents. Our investigation solely takes into account the static features, namely the prevailing weather conditions and pertinent road information. The ultimate goal is to establish a straightforward, yet comprehensive and integrated prediction model, leveraging a dataset that comprises in excess of one million documented incidents.

In addressing the prediction of accident duration, a bimodal approach was first implemented to categorize accident durations into short-term and long-term classifications. The random forest classifier emerged as the superior model in terms of evaluation metrics. Subsequently, the durations of both short-term and long-term accidents were predicted separately using various regression models where LightGBM model proved most effective in terms of MAE and RMSE.

Observing that this study is based on a large dataset containing more than a million incidents provides substantial evidence for the validity of our models. A practicable pipeline, combining the optimal classification and regression models, was thus constructed. This integrated model was capable of predicting traffic accident duration reliably while only using weather and road condition information, forming a comprehensive and reliable system for predicting accident durations. The simplicity and thoroughness of our methodology ensure a reliable and accurate prediction system that can be applied in real-world scenarios for traffic accident duration estimation, thereby contributing to improved traffic management and emergency response strategies in Texas and regions with similar weather conditions, such as Bangladesh.

**Code Sharing**

The complete work is performed and shared in Kaggle notebooks. The resources can be found here

https://www.kaggle.com/rafattabassumsukonna/a-bi-level-framework-for-traffic-accident-duration

**Declaration of generative AI and AI-assisted technologies in the writing process**

During the preparation of this work the authors used GPT-4 model within ChatGPT web application provided by OpenAI in order to check grammatical mistakes and improve the overall quality of the writing in sections they felt necessary. After using this tool/service, the authors reviewed and edited the content as needed and take full responsibility for the content of the publication.

**Declaration of Statement**

The authors declare no conflict of interest or competing interest in performing this work.

**References**

[1]  D. Mfinanga, E. Fungo, Impact of incidents on traffic congestion in Dar es Salaam city, Int. J. Transp. Sci. Technol. 2 (2013) 95–108.